\documentclass[envcountsect]{llncs}
\usepackage{graphicx}
\usepackage{amsmath,amssymb} 
\usepackage{color}
\usepackage[width=122mm,left=12mm,paperwidth=146mm,height=193mm,top=12mm,paperheight=217mm]{geometry}

\usepackage{multirow}
\usepackage{makecell}
\usepackage{amsmath}
\usepackage{fixltx2e}
\usepackage{capt-of,etoolbox}
\usepackage{silence}
\DeclareMathOperator*{\argmax}{arg\,max}

\begin{document}

\pagestyle{headings}
\mainmatter

\title{Outline Objects using Deep Reinforcement Learning} 

\author{Zhenxin Wang, Sayan Sarcar\\
\{wang.zhenxin; sarcar.sayan\}@kochi-tech.ac.jp
}
\institute{Kochi University of Technology}

\maketitle

\begin{abstract}
Image segmentation needs both local boundary position information and global object context information.  
The performance of the recent state-of-the-art method, fully convolutional networks, reaches a bottleneck due to the neural network limit after balancing between the two types of information simultaneously in an end-to-end training style. 
To overcome this problem, we divide the semantic image segmentation into temporal subtasks.
First, we find a possible pixel position of some object boundary; then trace the boundary at steps within a limited length until the whole object is outlined. 
We present the first deep reinforcement learning approach to semantic image segmentation, called DeepOutline, 
which outperforms other algorithms in Coco detection leaderboard in the middle and large size person category in Coco val2017 dataset. 
Meanwhile, it provides an insight into a divide and conquer way by reinforcement learning on computer vision problems.

\end{abstract}

\section{Introduction}
Semantic image segmentation is a crucial and fundamental task for human being and higher animals, like mammals, birds, or reptiles, to survive \cite{shettleworth2010cognition}.
It plays an irreplaceable role in the visual system in object localization, boundary distinguishing at pixel level.
We intensively differentiate objects from each other in our everyday life and consider it as an easy and intuitive task.
However, semantic image segmentation is still a challenging and not fully solved problem in the computer vision field. 

With the breakthrough of deep learning \cite{lecun2015deep} in the image recognition filed \cite{NIPS2012_4824}, many researchers began to apply it to their studies \cite{silver2017mastering,mnih2015human,resnet_he2016deep}, making significant progress in the entire computer vision and machine learning community. 
However, in semantic image segmentation or other dense prediction tasks in computer vision, the degree of improvement did not meet that of image classification or object tracking.
Indeed, the former task needs both global abstraction of the object and local boundary information.
On the contrary, the Deep Convolutional Neural Network (DCNN) is successful in extracting features of different classes of objects, but it losses the local spatial information of where the border between itself and the background is.

\begin{figure}
\centering
\includegraphics[width=1\linewidth]{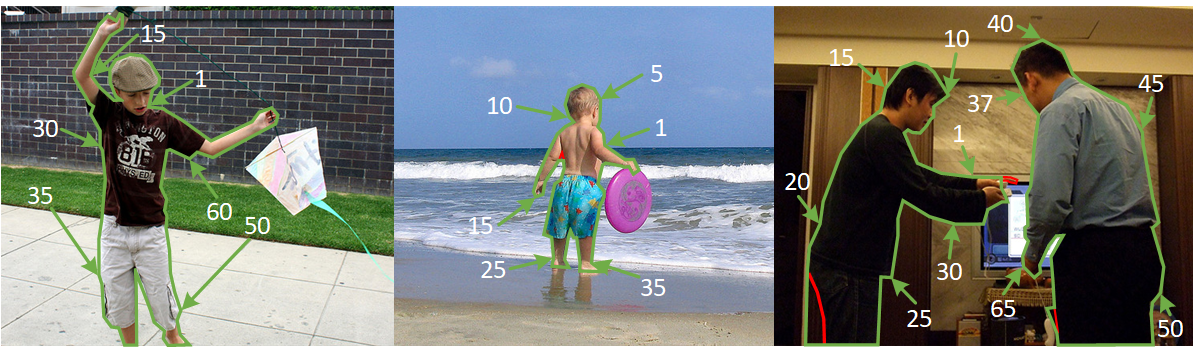}
\caption{Example outputs of the deep reinforcement learning agent -- DeepOutline, which sequentially generates polygon vertices of the outline of the recognizable object in the image. The green polylines show the final results of our outline agent of person category, while the red lines show the errors. The numbers in the image are the indexes of vertices in the generating order starting from 1.}
\label{fig:first}
\end{figure}

Many works have been done using an end-to-end dense predictor to direct predict the pixel label through score map for each class. 
But it is more challenging to combine the two diverse goals into one end-to-end deep neural network: local spatial information lost for the layers near the output label layer, especially for some extreme case, like different objects with similar texture, color, and brightness. 
Even when human segment these objects from an image, they need to keep the object classes in mind and distinguish around adjoining parts between objects to find out and trace the boundary gradually.
Some researchers seek to traditional computer vision method to overcome this problem, like Conditional Random Field (CRF) or hand-crafted features. 
But they need more sophisticated design than deep learning.

Such dilemma stems from the temporal combination of spatial information and global context. 
To solve this problem, we develop an end-to-end deep reinforcement learning network, called \textit{DeepOutline}, dividing segmentation task into subtasks (Figure \ref{fig:first} shows some successful results of DeepOutline).
Some of the questions we are trying to answer in this work are: 
Is it possible to imitate how human segment an image by outlining the semantic objects one by one?  
If we divide the segmentation task into steps, especially for the operation of finding the boundaries and finding the start point are quite different, can they share the same network?
It is easy to represent discrete actions, but how can we represent continuous outline position in the image?
Along with answering these questions, the contributions of this paper are threefold:

\begin{enumerate}
\item We present an end-to-end deep reinforcement learning network for semantic image segmentation, which copies a user holding a pen to draw the outline of objects in the image.  
\item We develop a \textit{position map} as the parameter to the agent action, and propose a \textit{Delta-net} to capture precise local spatial information combined with global context information.
\item We give insight into solving computer vision problems by a divide and conquer strategy, in which operations and the parameters of the operation are separately presented by different parts of a network.
\end{enumerate}

Apparently, using CRF technique after or inside the action could improve the final result. 
But we are interested in how reinforcement learning can be applied to image segmentation like tasks.
So we prefer to use as easy action as we can rather than pursue a leaderboard position in this paper.

\section{Related work}

\subsection{FCN related work}

First, much of the current literature pays particular attention to fully connected DCNN \cite{FCN_Long_2015_CVPR}.
When images are inputted to a neural network, both local boundary and global context information flow through the network. 
Due to the local information becoming vague with the downsampling operation, it is very intuitive to use the layers without or with less downsampling to directly pass the local information in the network for the final dense prediction output. 
Many studies have been done in this perspective, from U-net \cite{unet_ronneberger2015u}, which kept more precise position information before each downsample pooling and combined them with the more global context information passing back from the higher layers by upsampling operation, to the most recent fully convolutional DenseNets \cite{DenseNetSeg_JegouDVRB16}, which skipped connections between dense blocks \cite{Huang_2017_CVPR_DenseNet}. 
While deconvolutional layer \cite{FCN_Long_2015_CVPR} and unpooling \cite{unpooling_zeiler2010deconvolutional} tried to keep local position information by, respectively, transpose convolution and upsampling using previous pooling index to restore pooling context. 
Hyper column \cite{HyperColumn_Hariharan_2015_CVPR} was proposed to represent a pixel with deep features whose receptive field include it.
Other approaches were explored on improving the performance of image segmentation methods, like Convolutional Feature Masking \cite{dai2015convolutional}, Multi-task Netweork Cascades \cite{dai2016instance}, Polygon-RNN\cite{castrejon2017annotating}, and SDS \cite{SDS_hariharan2014simultaneous}.

To improve the boundary accuracy, another approach focuses on putting post process after labeling by DCNN, including CRF or even most sophisticated hand-crafted features. 
CRFs were mainly used in the semantic segmentation field to combine multi-way classifier scores with the low-level information captured by the local interactions of pixels and edges \cite{rother2004grabcut,shotton2009textonboost} or superpixels \cite{lucchi2011spatial}. 
Chen et al. \cite{deeplab_ChenPK0Y16} combined the responses at the final DCNN layer with a fully connected Conditional Random Field (CRF) \cite{krahenbuhl2011efficient}, to further improve the localization performance.

\subsection{Deep reinforcement learning}
Not long after Deep Mind got the first success in using Deep Reinforcement Learning (DRL) to achieve human level players skill in playing video games \cite{mnih2015human} in 2015, DRL got a major improvement, like winning the best human go player \cite{silver2016mastering}, spreading quickly to other field, like visual navigation \cite{zhu2017target}, Video face recognition \cite{rao2017attention}, visual tracking \cite{yoo2017action}, active object localization \cite{caicedo2015active}, image captioning \cite{ren2017deep}, joint active object localization \cite{kong2017collaborative}, semantic parsing of large scale 3D point cloud \cite{liu20173dcnn}, understanding game policies at runtime (visually-grounded dialoging) \cite{das2017learning}, attention-aware face hallucination \cite{cao2017attention}. 
Existing works of DRL for image segmentation are less focusing on understanding the actions and temporal correlation between them. In this work, we alternatively explore `Divide and Conquer' strategy in order to observe the characteristics of every atomic action human pursue for image segmentation through their eyes and the temporal correlation among them.  

\begin{table}[t]
\begin{center}
\caption{Terms of actions and states}
\label{state-action-terms}
\begin{tabular}{clcl}
\Xhline{2\arrayrulewidth}
Category                                                               & Term name    & Paramter  & \multicolumn{1}{c}{Discription}                                    \\ \Xhline{2\arrayrulewidth}
\multirow{3}{*}{Action}                                                & pen-up       & -         & To finish a polygon                     \\ \cline{2-4} 
                                                                       & pen-down     & Position  & To start a polygon or to draw to a position
                                                                       \\ \cline{2-4} 
                                                                       & draw-finish  & -         & To finish all polygons                  \\ \hline
\multirow{3}{*}{State}                                                 & pen-up       & -         & States after pen-up actions or before any action             \\ \cline{2-4} 
                                                                       & pen-down     & -         & State after pen-down actions                                    \\ \cline{2-4} 
                                                                       & draw-finish  & -         & The terminal state after pen-finish action                         \\ \hline
\multirow{2}{*}{\begin{tabular}[c]{@{}c@{}}Names\\ Alias\end{tabular}} & non-pen-down & -         & Action or state of pen-up or draw-finish                           \\ \cline{2-4} 
                                                                       & non-finish   & -         & Action or state of pen-up or pen-down                              \\ \hline
\end{tabular}
\end{center}
\end{table}

\section{DeepOutline}
This paper casts the problem of image segmentation as a reinforcement learning process, where a human-like agent finds out object boundaries one after another.
DeepOutline, holding a "pen", draws the boundaries of the objects in the input image, where the input image is treated as the environment of the agent. 
After it finds out all the recognizable objects in the image, it presses the stop button. 
In particular, we treat each object as one or more polygons combinations. 
The goal of DeepOutline is to find out all the polygons by sequentially generating the polygon vertices. 
The objective of this study is to investigate if dividing dense prediction task to subtasks (actions of agent) is possible, so we design the DeepOutline agent in an as simple form as it can segment images as below.

\subsection{Actions and state maps}
DeepOutline has two simple discrete actions, \textit{pen-up} and \textit{draw-finish}, and one complex action, \textit{pen-down}, which consists of a discrete action and the position parameter $\theta$ for the action. (We will introduce how we compute $\theta$ later in Section 3.3)
Correspondingly, DeepOutline has three states with the same name specifying the state after the action.
Actions, states and their related terms referred in this paper are listed in Table \ref{state-action-terms}

DeepOutline needs to keep the three kinds of data to show its states. 
The first data is the found outlines of object. The second is the current drawing outline, while the third is the last pen-down position. 
In our implementation, we use a matrix, called state map, of the same size of the input image to keep them.
To save the memory, the first two states are kept in one state map, and the last pen-down position takes one state map itself.

The default value of the state map is 0.
In the first state map, the found object polygons are drawn and filled with value 0.5, while the current unfinished polygon is drawn with value 1 to differentiate with the filled polygons even they sometimes overlap each other. 
This is because, sometimes, in the training data, the labeled regions of two objects next to each other may overlap; and sometimes one object is in front of another object in the image.
In the second state map, only the entry at last pen-down position is set to 1. 
If the last action is pen-up, then all the entries in the second state map are 0.

\subsection{Reward function}
DeepOutline has two types of reward functions. 
They separately measure how good the boundaries match when drawing (at each pen-down step) and how good regions match after each time DeepOutline finish a polygon drawing (at each pen-up step, sometimes a draw-finish directly following a pen-down also counts in this type reward).

The first type is the contour reward. 
It is calculated based on the closest match between boundary points in the ground truth and  segmentation maps outputted by DeepOutline.
In particular, we apply truncated Gaussian blur to the ground truth segmentation boundary map and calculate the sum of values of outputted boundary points in the blurred image. 
Let $B(x,y)$ be the blurred ground truth segmentation boundary map.
And let $Pt=\{(x_i,y_i)|(i=1,\cdots,n)\}$ be the new points added along the polygon after the action $a$, transferring the agent state from $s$ to $s'$.
Then the counter reward is defined as:
\begin{equation}\label{eq:3}
R_a^{contour}(s,s')=\frac{\sum_{i=1}^{n}B(x_i,y_i)}{\alpha} ,
\end{equation}
where $\alpha$ is a constant factor balancing counter and region rewards.

The second type is the region reward. 
We use normalized Intersection over Union (IoU) through all polygons in the training image as the reward score. 
Let $g$ be the ground truth polygon for the target object, and $p$ its segmentation polygon outputted by DeepOutline. 
Then the IoU between $p$ and $g$ is defined as:
\begin{equation}\label{eq:1}
IoU(g,p)=area(g\cap p)/area(g\cup p).
\end{equation}
Let $P=\{p_i|(i=1,\cdots,m)\}$ be the set of all found polygons, and $G=\{g_i|(i=1,\cdots,n)\}$ the set of all the ground truth polygons in the image, where $n\geq m$, and each $p_i$ is corresponding to $g_i$.
Then the region reward function after each non-pen-down action following a pen-down action is defined as 
\begin{equation}\label{eq:2}
R_{a_{non-pen-down}}^{region}(s_{pen-down},s')=\sum_{i=1}^{m} \frac{IoU(g_i,p_i)}{n} .
\end{equation}
After each action, the total reward is the sum of all possible rewards of the two types. But for continuous non-pen-down actions, the negative reward -1 is given. (See Table \ref{reward})
\setlength{\tabcolsep}{4pt}
\begin{table}
\begin{center}
\caption{Reward functions}
\label{reward}
\begin{tabular}{lll}
\Xhline{2\arrayrulewidth}
State                     & Action      & Reward                   \\ \Xhline{2\arrayrulewidth}
\multirow{3}{*}{pen-up}   & pen-up      & -1                       \\ \cline {2-3}
                          & pen-down    & $r^{contour}$            \\ \cline {2-3}
                          & draw-finish & -1                       \\ \hline
\multirow{3}{*}{pen-down} & pen-up      & $r^{contour}+r^{region}$ \\ \cline {2-3}
                          & pen-down    & $r^{contour}$            \\ \cline {2-3}
                          & draw-finish & $r^{contour}+r^{region}$ \\ \hline
\end{tabular}
\end{center}
\end{table}
\setlength{\tabcolsep}{4pt}

\begin{figure}
\begin{center}
   \includegraphics[width=0.7\linewidth]{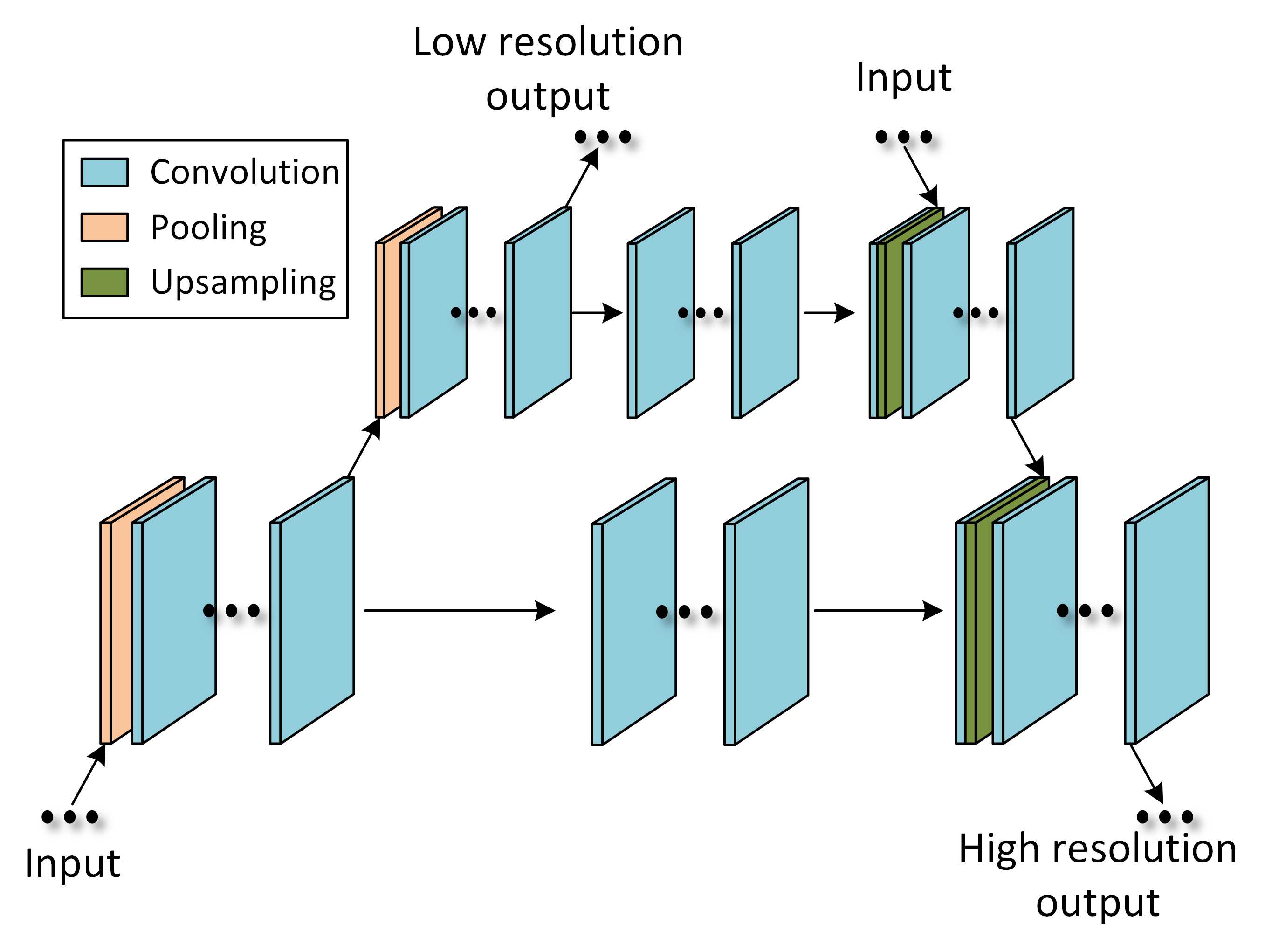}
\end{center}
   \caption{A Delta-net structure of 2 layers. It can easily be stacked with more layers. 
            Note that, the first input layer (down-left) of stacked Delta-net does not have the 
            pooling, and the last layer ignore the lower-resolution input (up-right).  }
\label{fig:deltanet}
\end{figure}

\begin{figure}
\begin{center}
   \includegraphics[width=0.4\linewidth]{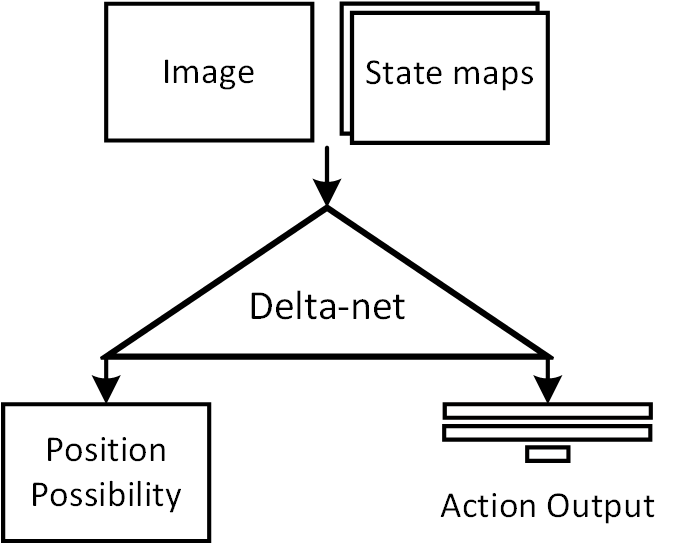}
\end{center}
   \caption{DeepOutline network structure and data flow chart.}
\label{fig:do-structure}
\end{figure}

\setlength{\tabcolsep}{4pt}
\begin{table}
\begin{center}
\caption{DeepOutline network architecture. All the convolution layers of the model use the same kernel size 3 with stride 1 followed by ReLU activation. There are max pooling operations of $2\times2$ between every next convolution blocks. conv$m-n$ stands for convolutional layers kernel size $m$ and outputs $n$ feature maps, $l$ stands for layer repeat times}
\label{Network-Architecture}
\begin{tabular}{lll}
\Xhline{2\arrayrulewidth}
Downward                                                                           & Horizontal & Upward                                                                                 \\ \Xhline{2\arrayrulewidth}
\begin{tabular}[c]{@{}l@{}}\textbf{Input} image\\ and state map\\ $(640\times640\times5)$ \end{tabular} &            & \begin{tabular}[c]{@{}l@{}}\textbf{Output} position\\ possiblility map\\ $(640\times640)$\end{tabular} \\ \hline

\begin{tabular}[c]{@{}l@{}}$\downarrow$ conv3-96 l=2\\$\downarrow$ conv3-96 $\quad\:\:\rightarrow$ \end{tabular} & \begin{tabular}[c]{@{}l@{}} \\ conv3-48 l=3 $\rightarrow$\end{tabular} & \begin{tabular}[c]{@{}l@{}} conv3-1  $\quad\quad\uparrow$  \\ conv3-48 l=3 $\uparrow$\end{tabular} \\ \hline
\begin{tabular}[c]{@{}l@{}}$\downarrow$ max pooling\\ $\downarrow$ conv3-128 l=2\\ $\downarrow$ conv3-128 \quad $\rightarrow$\end{tabular} & \begin{tabular}[c]{@{}l@{}} \\ \\ conv3-48 l=3 $\rightarrow$\end{tabular} & \begin{tabular}[c]{@{}l@{}} \\ \\ conv3-48 l=3 $\uparrow$\end{tabular} \\ \hline
\begin{tabular}[c]{@{}l@{}}$\downarrow$ max pooling\\ $\downarrow$ conv3-128 l=2\\ $\downarrow$ conv3-128 \quad $\rightarrow$\end{tabular} & \begin{tabular}[c]{@{}l@{}} \\ \\ conv3-48 l=3 $\rightarrow$\end{tabular} & \begin{tabular}[c]{@{}l@{}} \\ \\ conv3-48 l=3 $\uparrow$\end{tabular} \\ \hline
\begin{tabular}[c]{@{}l@{}}$\downarrow$ max pooling\\ $\downarrow$ conv3-256 l=2\\ $\downarrow$ conv3-256 \quad $\rightarrow$\end{tabular} & \begin{tabular}[c]{@{}l@{}} \\ \\ conv3-48 l=3 $\rightarrow$\end{tabular} & \begin{tabular}[c]{@{}l@{}} \\ \\ conv3-48 l=3 $\uparrow$\end{tabular} \\ \hline
\begin{tabular}[c]{@{}l@{}}$\downarrow$ max pooling\\ $\downarrow$ conv3-256 l=2\\ $\downarrow$ conv3-256 \quad $\rightarrow$\end{tabular} & \begin{tabular}[c]{@{}l@{}} \\ \\ conv3-48 l=3 $\rightarrow$\end{tabular} & \begin{tabular}[c]{@{}l@{}} \\ \\ conv3-48 l=3 $\uparrow$\end{tabular} \\ \hline
\begin{tabular}[c]{@{}l@{}}$\downarrow$ max pooling\\ $\downarrow$ conv3-512 l=2\\ $\downarrow$ conv3-512 \quad $\rightarrow$\end{tabular} & \begin{tabular}[c]{@{}l@{}} \\ \\ conv3-48 l=3 $\rightarrow$\end{tabular} & \begin{tabular}[c]{@{}l@{}} \\ \\ conv3-48 l=3 $\uparrow$\end{tabular} \\ \hline
\begin{tabular}[c]{@{}l@{}}$\downarrow$ max pooling\\ $\downarrow$ conv3-512 l=2\\ $\downarrow$ conv3-512 \quad $\rightarrow$\end{tabular} & \begin{tabular}[c]{@{}l@{}} \\ \\ conv3-48 l=3 $\rightarrow$\end{tabular} & \begin{tabular}[c]{@{}l@{}} \\ \\ conv3-48 l=3 $\uparrow$\end{tabular} \\ \hline
\begin{tabular}[c]{@{}l@{}}$\downarrow$ FC-128\\$\downarrow$ FC-128\\$\downarrow$ FC-3(\textbf{Output})\\ Soft-max\end{tabular}          &            &                                                                                        \\ \hline
\end{tabular}
\end{center}
\end{table}
\setlength{\tabcolsep}{4pt}

\subsection{Network architecture and Delta-net}
Many previous image and video related DRLs \cite{mnih2015human,silver2016mastering} use DCNN architecture, taking one or several images as input, followed with stacks of convolutional and pooling layers, then fully connected layers, and outputting action choices. 
Usually, in these studies, the agent states are implicitly included in the input image and the actions are treated to be discrete without any parameter.
Due to the different settings of this work, we make two changes to the original DRL structure.
First, our agent states are explicitly preserved in the state maps, then inputted to the DRL network together with the image.
Second, the pen-down action has a position parameter. 
We use a \textit{position map}  of the same size of the input image to estimate the possibilities of the pen-down position. Note that it is not a distribution in which the sum of all positions are 1. 
For the first pen-down for each polygon, the position $\theta = \argmax {(position map)}$,
while position $\theta = \argmax {(position map_{neighbor})}$ for other pen-down actions, where ${positionmap_{neighbor}}$ is the neighbor of $\theta$ of the last pen-down action.
For the non-pen-down actions, the similar network architecture to previous DRLs \cite{mnih2015human} is used.

To learn a position map at each pen-down action, we face the same problem to the DCNN-based image segmentation.
The difference is that we need the precise boundary positions iteratively outputted by the network, but not a dense prediction map.
We got inspiration from U-net \cite{unet_ronneberger2015u}, which keeps local information and concatenate them to upsampling layer passed back from low-resolution convolutional layers. 
However, when it is used for position map estimation, it does produce good estimation result.
It is partly because the convolutional layers from high to low resolution simultaneously keep both global context and local position information. 
We propose a stackable network substructure, called \textit{Delta-net}, by adding convolutional block before it concatenates to upsampling layer to keep and process the local position information (See Figure \ref{fig:deltanet}).
Delta-net has one input and two outputs in high and low resolution. 
The low-resolution output captures global context feature, while the high-resolution output can preserve local spacial feature combined with global information. 
Note that the added convolutional block can be in any form, including DenseNet \cite{Huang_2017_CVPR_DenseNet} or ResNet \cite{resnet_he2016deep}, etc.
It has a similar structure as Top-Down Modulation in (TDM)\cite{sameasme_ShrivastavaSMG16}. But they are different in both structures and functions. 
TDM only has one output in the middle-resolution end for ROI proposal and classifier while Delta-net has two outputs for action prediction in the low-resolution end and position prediction in the high-resolution end separately. 

\section{Experiments}
In our experiment, DeepOutline is built with 7-block Delta-net, taking the RGB image and agent state maps as input. 
It convolutes the feature of the highest resolution to 1 single map as the position map. 
Similar to traditional DCNN, the lowest resolution feature is followed with fully connected layers to output the action possibility (See Figure \ref{fig:do-structure} for the brief network structure and Table \ref{Network-Architecture} for the details). 
The loss function of action output is the sum of cross-entropy, and the loss of the position map is mean square. 
We weight them to a total loss and apply RMSProp with the learning rate of 1e-6, and other parameters are set as TensorFlow defaults.
The mini-batch size is set to 5 due to the GPU memory limitation. 
For the agent reward function, we set the windows size to 9, sigma 2 for the truncated Gaussian blur, and set the balance $\alpha$ to 0.05.

We train and test the proposed method on Microsoft Coco \cite{coco_lin2014microsoft} dataset. 
First, most of the segmentation annotations in the dataset are given by polygons, which facilitates DeepOutline to calculate the counter rewards. 
The second advantage of the Coco dataset is that it has a big number of segmentation instances. 
However, the amounts of instances in different classes are not balanced. 
There are 257k person polygon segmentation instances in the Coco train2017 dataset, but only 225 toasters. 
Unlike other categories of more than 10k instances, the person category has the most variety of deformation, viewing angles, occlusion, and sizes. 
So we train our model with the non-crowd person segmentation annotations in Coco train2017 and test it by Coco val2017.

\subsection{Man-made training data}
The exploration time could be very long and the exploration history needs much space because there are a large number of pen-down positions, which can make exponential increase of action sequences. To solve this problem we use an combined training data set by both exploration and  training example generated by human Coco segmentation annotations data. 
To simplify the procedure, the data is only generated by the polygon segmentation annotation in two kinds: 
The first is from pen-up state to pen-down action, the position map has a high value when the position is near the boundary of an unrecognized object. 
In particular, we first draw the polygons of unrecognized objects to the position map, apply the truncated Gaussian blur of 9 by 9 window and the sigma of 2 (blur step), and then scale the map to the range between 0 and 1 (scale step). 
The second is from pen-down state to pen-down action. 
This kind of position map represents the possibility of next vertex position. We set the positions near the polylines to be drawn within 50 pixels in length to higher values.
There is a balance of choosing between fewer polygon vertices to get higher reward within fewer actions and more vertices to get a higher boundary fit, but the only information drawing by the Coco dataset to us is that human segmentation annotations are not so precise.
So we assume the training dataset as a coarse dataset and use a later reinforcement learning exploration to refine the training.
Let $\mathit{seg}_i=\{\mathit{p}_i^j=(x_j,y_j)|j=(1,\cdots,k,\cdots)\} \in SEG $ be the current polygon segment and DeepOutline has picked up the $1,\cdots,k$ vertices, where $SEG$ is the segmentation annotations for image I in Coco dataset. 
We first set the value of the position along $\mathit{seg}_i$ from vertex $\mathit{p}_i^k$ within length of 50 to the following piecewise function of the length to vertex $k$ along $\mathit{seg}_i$:
\begin{equation}\label{eq:4} 
f(p)= 
 \begin{cases}
   \frac{|p-\mathit{p}_i^k|}{|\mathit{p}_i^{k+1} -\mathit{p}_i^k|} &\text{if}\:p\: \text{is on the line} \\
     & \quad\quad\quad(\mathit{p}_i^k,\mathit{p}_i^{k+1})\\
    1-\frac{|p-\mathit{p}_i^{k+1}|}{50-|\mathit{p}_i^{k+1} -\mathit{p}_i^k|}&\text{if}\:p\: \text{is on the polyline} \\      
     & \quad\quad\quad(\mathit{p}_i^{k+1},\cdots)\\
 \end{cases}
\end{equation}
where $p$ is the point variable in the position map, $|\cdot|$ is the euclidean distance along $\mathit{seg}_i$
Then we apply the same blur and scale steps with the same parameters in the first kind of man-made data.

\subsection{Agent exploration}
Each image has a training history 2k-length queue for saving the explored states and Q-values.
Initially, the history queue is filled with man-made training data.
DeepOutline is trained in phases.
In each phase DeepOutline randomly chooses 10 images, loads their history queue,  explores and trains the network with $\epsilon$-greedy policy for 500 steps. 
The explored possible action space and Q-values are put to the history queue which is saved to the disk for future training phases.
We set the initial exploration $\epsilon$ to 0.8 and linearly scale it down to 0.05 in 10m training steps.

Finding the polygon boundary need not only the discrete action indicating it starting or finishing to draw the object but also the precise position each time it draws in the image. 
In our implication, we set the max length between two next pen-down action to 50 pixels.

\subsection{Equal possibility of training actions}
The network has three action outputs. 
Most of the outputs are pen-down, but less pen-up and draw-finish.
However, if the network is trained with unbalanced examples, i.e. too many pen-down comparing to pen-up or draw-finish, then the trained network is lean to produce more than it should, or even does not output any non-pen-down action.
In order to overcome this problem, we select our mini-training-batch with equal possibilities of the three actions.

\setlength{\tabcolsep}{4pt}
\begin{table}
\begin{center}
\caption{Segmentation results on person category of Coco val2017. AP\textsubscript{S}, AP\textsubscript{M}, and mAP\textsubscript{L} is the mean Average Precision for small (area $< 32^2$), medium ($96^2 >$ area $> 32^2$) and large (area $> 96^2$) objects respectively. Note that our result is tested on Coco val2017 while other results on the Coco detection leaderboard are tested on test2017. \color{red}The results would be update after iterations of training.}
\label{experiment_result}
\begin{tabular}{lllll}
\Xhline{2\arrayrulewidth}
                & mAP    & AP\textsubscript{S}   & AP\textsubscript{M}   & AP\textsubscript{L}   \\ \Xhline{2\arrayrulewidth}
UCenter \cite{liu2018path}         & \textbf{0.534} & \textbf{0.343} & 0.589 & 0.728 \\ \hline
Megvii (Face++) \cite{leaderborad} & 0.524 & 0.344 & 0.573 & 0.706 \\ \hline
FAIR Mask \cite{fair_mask_rnn_HeGDG17} & 0.496 & 0.293 & 0.555 & 0.698 \\ \hline
MSRA \cite{MSRA_DaiQXLZHW17}           & 0.491 & 0.317 & 0.543 & 0.669 \\ \hline
MSRA\_2016 \cite{MSRA2016_LiQDJW16}     & 0.43  & 0.219 & 0.520  & 0.645 \\ \hline
\textbf{DeepOutline}   & 0.530 & 0.297 & \textbf{0.631} & \textbf{0.730} \\ \hline
\end{tabular}
\end{center}
\end{table}
\setlength{\tabcolsep}{4pt}

\subsection{Experiment results}
Table \ref{experiment_result} shows the segmentation result of the proposed model on the person category of Coco val2017 set. 
Our model performed best for middle and large size objects. 
For small size objects, 
when DeepOutline traces along a limb to the end, it has a chance hardly to find out the way back and may chooses boundary of its neighbor, and make the segmentation inaccurate (See Figure \ref{fig:fail}a and b).

Figure \ref{fig:lastImg} shows more position maps on Coco val2017 set during the outline process. 
The first two rows show the maps at the beginning of outline, at the second pen-up state and at the pen-down state in outlining the second polygon. 
In pen-up states, the outer edges of person have higher value, while in the pen-down states, only the edges near the last pen-down position have a higher value. 
The third to the fifth row only show maps for two pen-up stats.
And one failure mode is shown at the last row.

\begin{figure}
\begin{center}
   \includegraphics[width=0.45\linewidth]{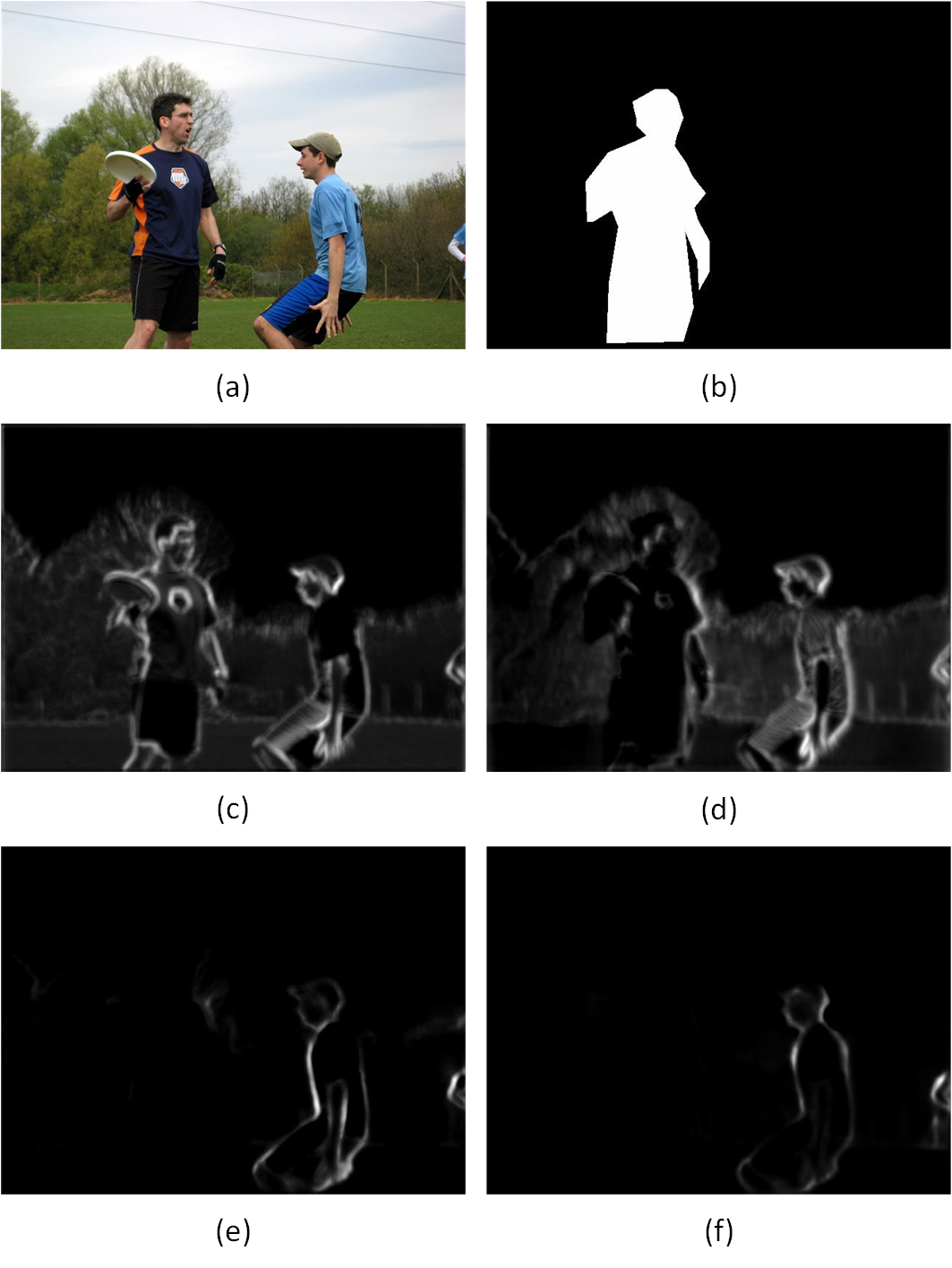}
\end{center}
   \caption{Evolution of position map during training. (a) is original image and (b) is the first state map of pen-up state generated by the training segmentation data. 
   So the second state map is 0 and not shown here. 
   (c) to (f) are maps of the first pen-down position of the second polygon at training round 20k, 40k, 60k and 100k respectively.}
\label{fig:porcess}
\end{figure}

\subsection{Further findings}
\begin{figure*}[t]
\begin{center}
   \includegraphics[width=1\linewidth]{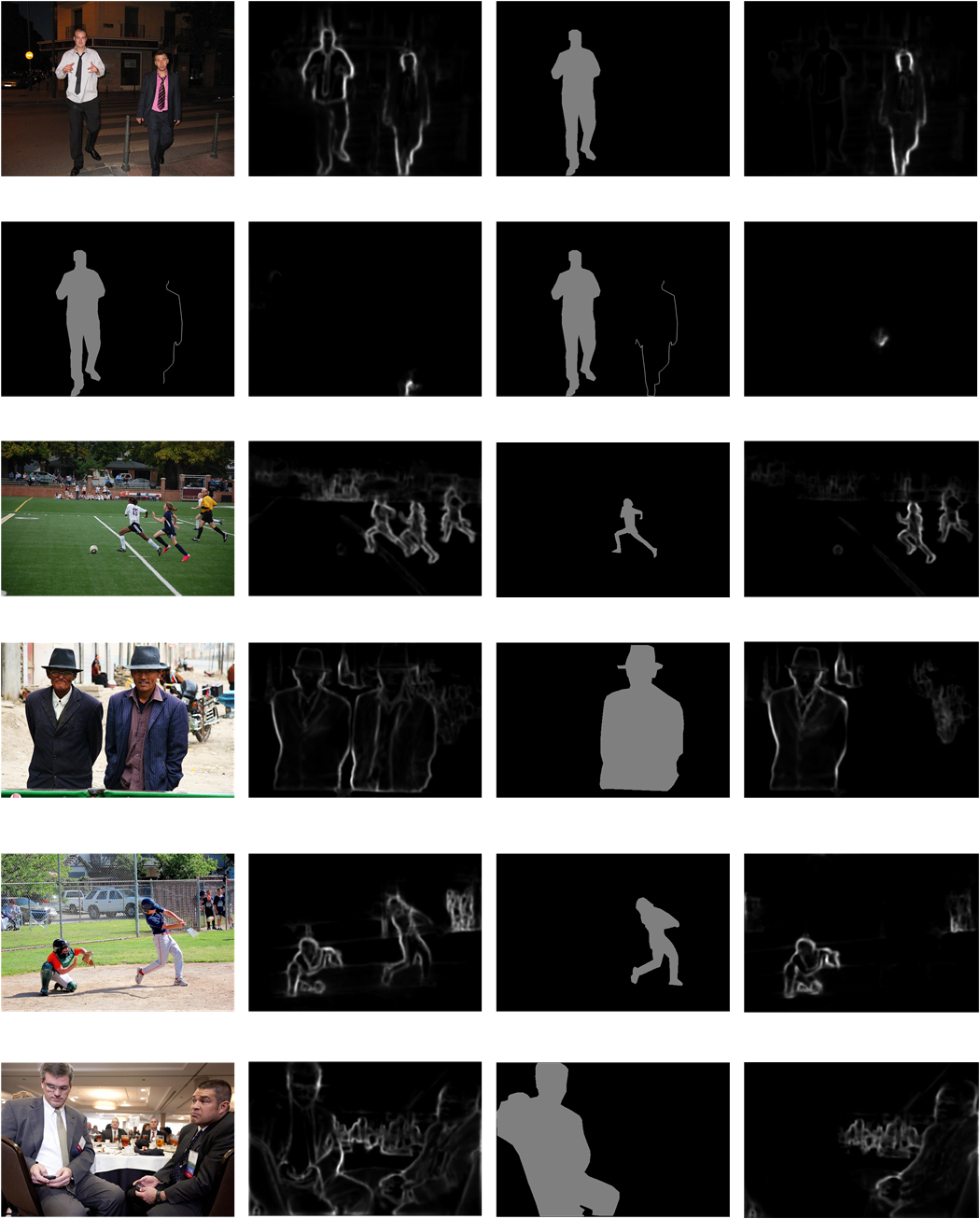}
\end{center}
   \caption{Position map visualization on Coco val2017 set. The first and the third columns are images and states map 1. The second row shows selected pen-down state steps. The last row shows one failure mode (hair on the shoulder).}
\label{fig:lastImg}
\end{figure*}

In Figure \ref{fig:lastImg}, when there are some finished segmentations in the state map, the corresponding regions in the position map are suppressed.
To understand how it happens, we traced the training process. 
Between 10k and 40k effective training steps in which the training data show the promising actions winning more positive reward, DeepOutline began to find out possible edges in the image but ignored the state map input.
At this time, the position outputs of DeepOutline showed no difference to the change of state maps, but between 30k and 80k effective training, the position map outputs started to suppress the interior edge response.
And after that, the network needed a long term to fit the position state map. 
We trained the network 4 times on 2 PCs. They all showed the same trends although the changes between phases were slow and they did not keep the same pace.

\subsection{Failure cases and limitations}

\begin{figure}
\begin{center}
   \includegraphics[width=0.45\linewidth]{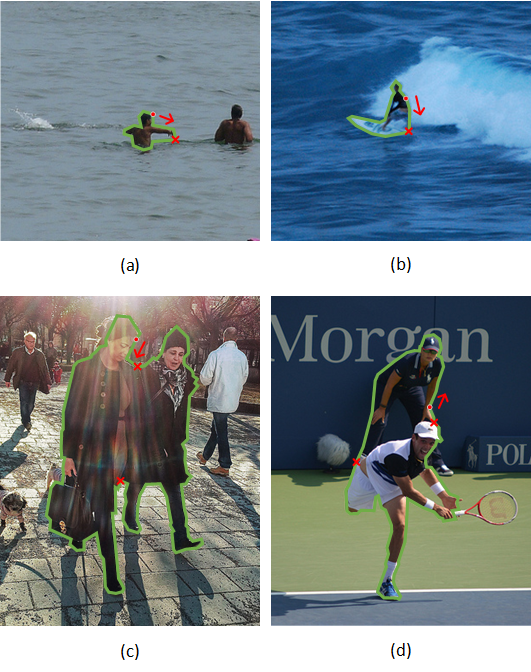}
\end{center}
   \caption{Failed examples in Coco val2017 dataset. 
   The red dot is the start point, and the arrow stands for tracing direction, and the red cross indicates the chosen vertex before the first wrong action.
   (a) and (b) are small targets. 
   The clothes of the two women in the middle are similar and without distinct boundary in between. 
   While (d) mixes body parts of two people.}
\label{fig:fail}
\end{figure}

Figure \ref{fig:fail} illustrates example of failure cases. 
Most of them are caused by some middle action wrongly taken by the agent network.
We identified three categories of wrong choices. 
The first type of errors took place for small targets (See Figure \ref{fig:fail}a and b).
The width of the right foot end of the person in Figure \ref{fig:fail}b is in less than three pixels in the original image. 
It is so small that the change of going along the surf board in the position map was higher than the right direction of tracing back along the right leg. 
This also happened for some small and spiked parts of the object, like the finger marked around red in the right image of Figure \ref{fig:first} and the sea wave below the armpit of the kid in the middle image in Figure \ref{fig:first}.
The second type example is shown in Figure \ref{fig:fail}c, which was caused by similar texture, fuzzy boundary or similar, especially in the case where two close objects share the same class. 
The third type is that the parts of the same class adjoin together even if they have clear boundary (See Figure \ref{fig:fail}d).

\begin{figure}[t]
\begin{center}
   \includegraphics[width=0.45\linewidth]{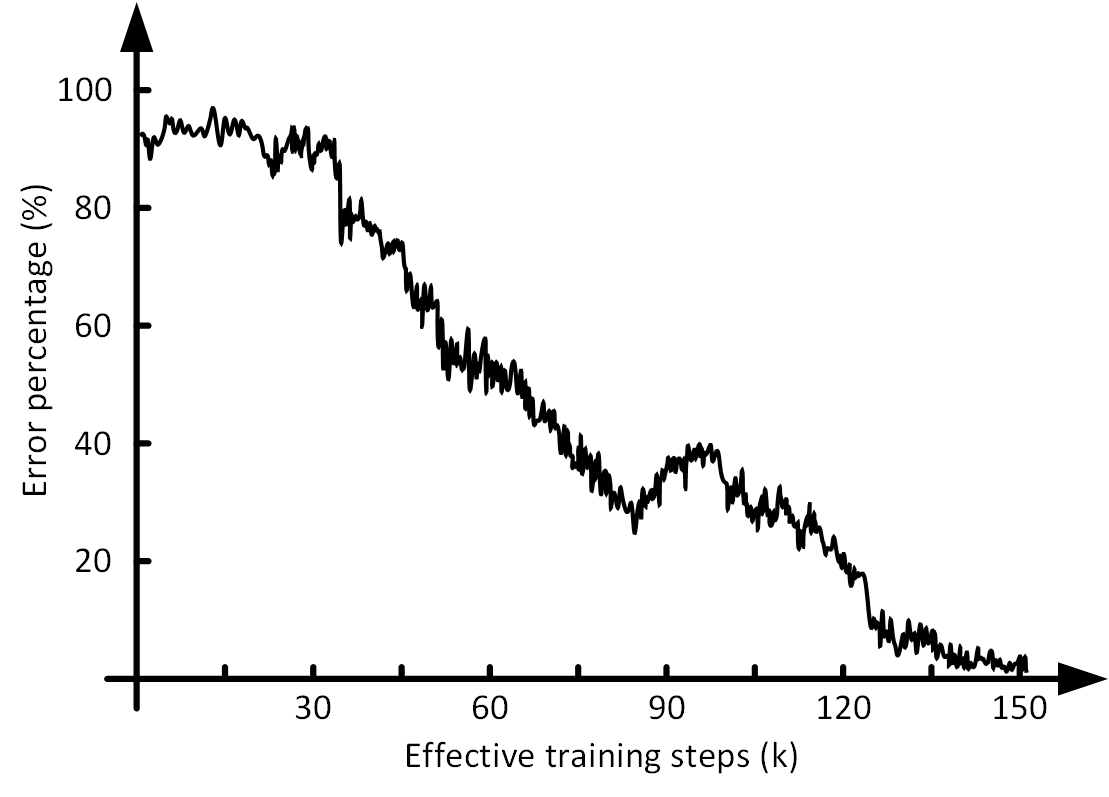}
\end{center}
   \caption{Percentage of wrong DeepOutline actions through the training stage.  }
\label{fig:error}
\end{figure}

When the segmentation task is divided into sequential subtasks, the requirement for the reliability of each subtask goes high. 
Fortunately, the percentage of wrong action outputs by DeepOutline decrease during the training stage showed in Figure \ref{fig:error}.
After 150k effective training rounds, the possibility of wrong action output reached below 2 percent.
Note that, not all the wrong actions would cause a failure, in which the agent outlines two different objects or generate pen-up action before it finish the polygon.
Many of them make a missing or a redundant part to object.
The performance under this kind of problems can be improved by developing complex agent actions with more errorproof mechanisms. 

There are also other reasons like the train data has various criterions on annotation by different annotators. 
Some of them treat the occlusions as a different object while others ignore them and only annotate the object. 
This also raise the chance of early stop action, non-stop and tracing a boundary of a different object.

\section{Conclusions}
In this paper, we showed the first deep reinforcement learning method for image segmentation. 
We designed DeepOutline with simple actions by outlining the semantic objects one by one, and it outperforms other algorithms in Coco detection leaderboard in the middle and large size person category in Coco val2017 dataset. 
We discussed how the network behaves under different state maps and the causes of the three types of error.
This paper also argued that the paradigm of divide and conquer is a promising approach to dense prediction problems.

\subsection{Acknowledgement}
We gratefully acknowledge the support of NVIDIA Corporation with the donation of the Titan Xp GPU used for this research.

\clearpage

\bibliographystyle{splncs}
\bibliography{egbib}
\end{document}